# Decision Aids for Adversarial Planning in Military Operations: Algorithms, Tools, and Turing-test-like Experimental Validation

*Alexander Kott*

*Ray Budd*

*Larry Ground*

*Lakshmi Rebbapragada*

*John Langston*

**Abstract**

Use of intelligent decision aids can help alleviate the challenges of planning complex operations. We describe integrated algorithms, and a tool capable of translating a high-level concept for a tactical military operation into a fully detailed, actionable plan, producing automatically (or with human guidance) plans with realistic degree of detail and of human-like quality. Tight interleaving of several algorithms -- planning, adversary estimates, scheduling, routing, attrition and consumption estimates -- comprise the computational approach of this tool. Although originally developed for Army large-unit operations, the technology is generic and also applies to a number of other domains, particularly in critical situations requiring detailed planning within a constrained period of time.  In this paper, we focus particularly on the engineering tradeoffs in the design of the tool. In an experimental evaluation, reminiscent of the Turing test, the tool's performance compared favorably with human planners.

**The Quest for New Processes and Tools**

The US Army[1] is exploring a significant computerization of the military planning process [1]: "...the Army must create fast new planning processes that establish a new division of labor between man and machine. … Decision aids will quickly offer suggestions and test alternative courses of actions."

The reasons for exploring potential benefits of such decision aids are multifaceted. The process of planning an Army operation remains relatively cumbersome, inflexible and slow. Life and death decisions are made by a relatively meager staff of perhaps 3-4 individuals, working with imperfect information under significant time constraints. Success typically depends on one's ability to synchronize the movement of units and the concentration of firepower and other effects at a precise moment in time.

The tools available to the planners are generally limited to rudimentary decision aids for analyzing the terrain of the battlefield and some office automation tools to help record decisions. The planners bring varying degrees of knowledge, experience and prejudices to the process. The planning process frequently involves disagreements on estimation of outcomes, enemy reactions, attrition and consumption of supplies. There is a fundamental complexity of synchronization and effective utilization of multiple heterogeneous assets performing numerous, inter-dependent, heterogeneous tasks.

The Course of Action Development and Evaluation Tool (CADET), is a tool for producing automatically (or with human guidance) the detailed tasks required to translate a basic concept into a fully formed, actionable plan, which is a key step in the military's standard decision making process. This step involves taking the proposed courses of action for the friendly forces, developed in a previous step and initially expressed as high-level concepts, and expanding them into the hundreds of supporting tasks required to accomplish the intended objective. Concurrently, the friendly course of action is tested

---

[1] Views expressed in this paper are those of the authors and do not necessarily reflect those of the U. S. Army or any agency of the U.S. government.

against the most likely and/or the most dangerous courses of action available to the enemy. The intent of this analysis, called *wargaming* in the military, is to produce an analytical baseline from which the commander can choose the best course of action [2]. The process is particularly challenging for a relatively large and complex unit such as a Division, where the actions of 12,000 soldiers and over 3,000 weapons must be coordinated to produce a desired effect. Wargaming a potential tactical course of action for such a large unit typically involves a staff of 3-4 persons with in-depth knowledge of both friendly and enemy tactics.2

The input for their effort comes usually from the unit Commander in the form of two doctrinally defined products: a Course of Action (COA) sketch (e.g., Fig. 1) and a Course of Action statement -- a high-level specification of the operation. In effect, such a sketch and statement comprise a set of high-level actions, goals, and sequencing, referring largely to movements and objectives of the friendly forces, e.g., "Task Force Arrow attacks along axis Bull to complete the destruction of the 2nd Red Battalion." Typically, the unit Commander will develop a minimum of three courses of action for consideration, distinguishable from one another in force composition and application, designation of main and supporting efforts or in utilization of terrain and resources. The human planner applies his or her knowledge of the art of war to the creation of the courses of action so that each may generally be assumed, on the surface, to be feasible.

With this input, working as a team for several hours (typically 2 to 8 hours), the members of the planning staff examine the elements of each friendly Course of Action (COA) in detail. The process involves planning and scheduling of the detailed tasks required to accomplish the specified COA; allocation of tasks to the diverse forces comprising the Division or the Brigade; assignment of suitable locations and routes; estimates of friendly and enemy battle losses (attrition); predictions of enemy actions or reactions, etc.

---

2 The military has formalized this process into the Military Decision Making Process or MDMP. The CADET tool focuses on the course of action analysis phase of the MDMP.

The outcome of the process is usually recorded in a synchronization matrix [2], a type of Gantt chart. Time periods constitute the columns. Functional classes of actions, such as the Battlefield Operating Systems, are the rows (see Fig. 2). Examples include maneuver, logistics, military intelligence, etc. The content of this plan, recorded largely in the matrix cells, includes the tasks and actions of the multiple subunits and assets of the friendly force; their objectives and manner of execution, expected timing, dependencies and synchronization; routes and locations; availability of supplies, combat losses, enemy situation and actions, etc.

Although the immediate product may look like a prescriptive schedule, the purpose of the COA analysis process is certainly not to impose a rigid script for the battle. Rather, the purpose is to ascertain the feasibility of the COA, to assess its likelihood of success against a particular enemy COA, and to identify the range of probable actions and the points of synchronization for participants. Comparing the results of the wargames for the possible courses of action, the commander selects the best COA, in terms of accomplishing his desired intent with most effective application of resources. The final product is an executable COA translated into a synchronized operational plan.

**How CADET is Used**

In this complex, difficult and time-consuming process, the Course of Action Development and Evaluation Tool (CADET) assists military planners by rapidly translating an initial, high-level COA into a detailed battle plan, and wargaming the plan to determine if it is feasible.  Working with the planner in a series of user/computer actions, the system details, resources, schedules, elaborates, and analyzes the COA.

In brief, the human planner defines the high-level COA via a user interface (e.g., Fig. 3) that enables him to enter the information comparable to the conventional COA sketch and statement (e.g., Fig. 1), which the COA-entry interface then transforms into an input to CADET proper, a collection of formal assertions and/or objects, including typically on the order of 2-20 high-level tasks. Fig. 4 outlines an input to CADET. Essentially an electronic representation of a traditional sketch-and-statement, it describes the friendly and enemy units, control measures (e.g., boundaries of areas), network of traversable terrain (e.g., mobility corridors), the high-level scheme of maneuver represented as a set of activities (including the estimates of enemy activities), and temporal relationships between the activities. Most entities have numerous attributes in addition to those shown here; they can take a default value or be specified explicitly by the human planner. This definition of the COA is transferred to CADET, which proceeds to expand this high-level specification into a detailed plan/schedule of the operation.

Within this expansion process, CADET decomposes friendly tasks into more detailed actions; determines the necessary supporting relations, allocates / schedules tasks to friendly assets; takes into account dependencies between tasks and availability of assets; estimates enemy actions and reactions; devises friendly counter-actions; and estimates paths of movements, timing requirements, force attrition and supply consumption. The resulting detailed, scheduled and wargamed plan often consists of up to 500 detailed actions with a wealth of supporting detail (e.g., Fig. 5a and 5b).

Having completed this process (largely automatically, in about 20 seconds on a mid-level laptop computer), CADET displays the results to the user (e.g., Fig. 6 and 7) as a synchronization matrix and/or as animated movements on the map-based interface. The user then reviews the results and may either change the original specification of the COA or directly edit the detailed plan. Once a satisfactory product is reached (typically within 5 to 30 minutes), the user utilizes it to present the analysis of the courses of action to the Commander, and to produce operational orders. While the tool can make no claims of producing an optimal plan (indeed, it is highly dependent on the quality of the COA sketch and statement which originated with the unit Commander), it does provide a strong measure of the plan's technical feasibility in a short amount of time. With the tool helping speed along the application of the *science of war*, the unit Commander has more time available to apply his brainpower to the *art of war*. General George S. Patton, Jr. recognized the need for speed over optimality when he asserted "*[A] good plan violently executed now is better than a perfect plan next week*" [17].

Recently, several efforts have utilized the planning capability of CADET. The Defense Advanced Research Projects Agency (DARPA) used CADET for its Command Post of the Future Program and the Army's Battle Command Battle Laboratories used it to complement the capabilities of course of action tools being developed at Fort Leavenworth and at Fort Huachuca.[3] At this time, CADET is apparently the first and so far the only tool that was demonstrated to generate Army battle plans with realistic degree of detail and completeness, for multiple battle operating systems, and for the large scale and scope associated with such large, complex organizations as an Army Division or a Brigade. In the related domain of small-unit operations, Tate [5] has described a very mature work.

In experimental evaluations (discussed later), a CADET-assisted planner performed dramatically faster than a conventional human-only planning staff, with comparable quality of planning products.

Although originally developed for Army large-unit operations, CADET is largely generic and can be applied to a broad range of tasks that require interleaving of planning, resource scheduling and spatial movements. Being a knowledge-based tool, CADET is adapted to a new application domain by changing its knowledge base. In particular, we have already built exploratory demonstrations for such tasks as intelligence collection using scouts and unmanned aerial vehicles; combat tasks of robotic forces; and responses to terrorism incidence in an urban environment.

---

[3] Battle Command Battle Lab-Leavenworth chose CADET as a key element for its Integrated Course of Action Critiquing and Evaluation System (ICCES) program [3]. Battle Command Battle Lab-Huachuca integrated CADET with a military intelligence system called All Source Analysis System-Light to provide a planner for intelligence assets and to wargame enemy courses of action against friendly courses of action.

**Key Requirements and Challenges of the Problem Domain**

The needs of the problem domain clearly do not allow one to focus a useful decision aid on a narrow slice of the problem, e.g., only planning, scheduling, or routing. Strong dependencies exist between these aspects of the overall problem.

For example, considering the coupling between planning and scheduling, the activities of a military unit may differ dramatically depending on whether the timing of its arrival to a location occurs merely minutes before or after the arrival of the adversary units. The activities (planning) and timing (scheduling) associated with a move of a unit can affect the suitable route, e.g., the unit may have to take a more circuitous route in order to avoid detection by the enemy unit expected to threaten an area during a certain time period. The route, in turn, affects both the timing and the activities performed by the unit, e.g., a unit may have to bypass an enemy unit, or to perform a passage of friendly lines, or to require a set of logistics-related activities to supply the unit enroute. The attrition (losses of personnel and equipment to enemy actions or to breakdowns) and consumption of supplies such as fuel and ammunition depend strongly on the activities and movements of the unit. In turn, attrition and consumption affect the suitability and feasibility of certain activities that a unit can perform. Obviously, the overall decision-making process must also include elements of adversarial reasoning such as determination of enemy actions and reactions to friendly actions. Interestingly, the presence of an adversary tends to exacerbate the multi-directional coupling that exists between the different aspects of the problem -- planning, scheduling, routing, adversary prediction, attrition and consumption estimation. The pervasive nature of the coupling suggests that a strong integration (unification) of all the associated computational processes is required [6, 7, 28].

Also significant is the breadth of coverage in terms of the functional classes of tasks that must be explored and planned by the decision aid. While maneuver tasks are central to the battle, other functions, such as logistics or military intelligence are interdependent with the maneuver tasks and must be all analyzed in close integration.

In spite of the complexity implicit in these multiple interdependent problem aspects, speed is extremely important. We envision that the user would approach a tool like CADET somewhat similarly to a spreadsheet: he would enter or modify key elements of a COA and then expect to see the recalculated details of the plan almost instantaneously. There can be multiple iterations of modifications and recalculations, and it is most reasonable for a user in field conditions to expect a fast response in the order of seconds. Such a performance must be achieved with a laptop computer, perhaps even a handheld computer.

Because of rapidly changing elements of tactics, often evolving as operations unfold, and the differences in styles and procedures of different units and commanders, it is also imperative to provide a decision aid with the means to modify its knowledge base literally in field conditions, by end user, non-programmer.

Given that CADET is most likely to be in a framework of a larger deployed system, with its own style and implementation of the user interface, it is important to make the decision aid largely independent of user interface assumptions. For the same reason it is important to provide convenient and flexible means for interfacing the decision-aid component with other systems, including the case when the two systems do not share a common ontology.

**Possible Approaches, Criteria and Tradeoffs**

Given the challenges outlined in the previous section, we adopted the following set of primary criteria for selecting the technical approach to solving the CADET problem.

(a) A suitable technical approach must integrate closely the subproblems of planning, resource scheduling and movement routing.

(b) The approach must be frugal in terms of the knowledge base required, while covering a broad range of actions and effects present in the domain.

(c) Adversarial reasoning must be an integral component of the technical approach.

(d)  It must comply with the intended concept of the application use: the system must provide automated battle planning within the constraints of user's high-level plan concept, and within the established doctrinal rules understandable to the user.

Armed with these criteria, let us evaluate some of the alternative approaches potentially applicable to CADET's problem.

Planning being the central theme of CADET's application, it is appropriate here to explore applicability of classical planning techniques and their extensions. In this broad field of approaches, the world is represented by a state model, actions lead from one state to another, and the objective of the problem solver is to find a sequence of actions that lead from the initial state to the goal state or states. A recent surge of advances in the field has been originated by the powerful Graphplan solver [19].

From the CADET's needs perspective, these approaches do not meet several criteria. First, the need for resource scheduling and movements routing within the planning process. Although there is an active area of research in temporal planning where the classical planning model is extended to account for temporal constraints, it does not cover all the issues involved with resource-constrained scheduling (e.g., [27]). Second, when applied to realistic problems, the state model becomes excessively large and difficult to build which leads for strong arguments that practical problems are best solved by knowledge-based planning rather than classical planning approaches (e.g.., [20]).

In our approach, as discussed later, CADET integrates closely resource allocation and scheduling within an interleaved planning-scheduling process, and uses a frugal knowledge representation typical of knowledge-based planning that does not require an explicit state and action model. CADET's approach comes with tradeoffs: it does not enjoy the theoretical rigor, optimality and provable properties characteristic of many approaches built on classical planning.

An example of a practical extension to classical planning is SIPE-2 [22] which avoids many of the limitations of classical planning by adopting a more flexible and potentially more frugal representational scheme. This approach also demonstrated successes in a loose integration with a resource scheduler OPIS [21]. Like CADET, development of SIPE-2 has been motivated by challenges of military planning problem.

In exploring the applicability of SIPE-2 to CADET problem, we note (a) SIPE-2's restricted and limited integration with scheduling, especially as may relate to the time-space routing requirements of CADET and (b) lack of provisions for integrated adversarial reasoning. In CADET, routing and resource scheduling are tightly integrated and adversarial reasoning is a part of the overall planning process.

If we were to adopt SIPE-2, we could have availed ourselves of its rich facilities and available integration with other components that support automated plan monitoring, reactive plan repair, extensive user interfaces and knowledge-base authoring tools. However, it was felt that these were less critical to CADET application then the requirements of tight integration of scheduling, routing, and adversarial actions.

Prominent among approaches with specific emphasis on adversarial actions is Adversarial Planner (AP) -- a state-based, application-independent, hierarchical-task-network planner [23]. Its adversarial reasoning process models the Army's action-reaction-counteraction COA planning protocol.

AP's problem solver determines adversary's likely counteraction by searching for actions that would negate a precondition for own action. Although originally developed for military applications, AP has been recently applied to control and robotic in space applications [24] where variety of adverse conditions and breakdowns are analogous to a military adversary.

Applicability of AP's approach to CADET faces concerns with respect to at least three criteria: (a) explicit encoding of actions pre- and post-conditions demands a rather expensive, state-based representation scheme; (b) resource scheduling is not an integral part of AP even though temporal planning is, and (c) it is doubtful that users would be comfortable with a solution process that automatically determines suitable counteractions in a manner that may not be transparent to the user. Although use of AP would have provided the CADET system with AP's inherent execution monitoring and reactive repair mechanisms, these were not critical considerations for the CADET application.

A different class of approaches to planning in adversarial environments is based on game-theoretic techniques. Examples include recent work in application to battle planning and control [25], [26]. The problem is formulated as a game and the goal is to determine the strategy that optimizes the objectives of the players. The resulting solution represents a plan of action for the players.

While  attractive in terms of explicit and rigorous treatment of problem's adversarial aspects, such game-theoretic approaches fail to meet CADET's criteria: (a) the  CADET solution must comply with the high-level strategy formulated by the user and  it is difficult to accomplish such a compliance within solution that automatically generates a strategy; (b) the cited approaches are effective for problems formulated in terms of spatial moves and weapon firings, while many of actions within a CADET are of different nature (e.g., coordination between units) that do not lend themselves readily to a board game representation. In declining the use of game-theoretic approaches there was a tradeoff: CADET does not have a significant capability to generate automatically a sophisticated temporal-spatial strategy of actions, and must rely on the user to define such a high-level strategy.

**The CADET Technical Approach**

Perhaps the most fundamental engineering choices were the basic functional focus and concept of user operation of the tool. CADET focuses on course of action analysis, which is the most time-consuming aspect of the Military Decision Making Process.  Other researchers (e.g., [8, 9, 10, 4]) are addressing a different (and preceding) phase of the process,  the very interesting and challenging problem of generating the high-level maneuver COA. In addressing the style of interactions between the human and the decision aid, we de-emphasize the mixed-initiative, incremental style (even though CADET allows such a style) in favor of a rapid style of generating a complete plan from a high-level COA, followed by manual modifications. Motivations include the desire to minimize the time-consuming aspects of a mixed-initiative process; to avoid significant dependencies on the style of user interfaces; and to eliminate the additional complexity of still poorly understood issues of mixed initiative in an already very complex problem.

*Integration of Multiple Problem Aspects*

The integration of multiple distinct problems within the overall process is achieved via an algorithm for tightly interleaved incremental planning, routing, time estimating, scheduling, estimates of attrition and consumption, and adversarial reaction estimation. This interleaving approach descends conceptually from [11] where similar interleaving applied to a design domain.

As shown in Fig. 8a, the hierarchical task network-like planning step produces an incremental group of tasks by applying domain-specific expansion rules (Fig. 8b) to those activities in the current state of the plan that require hierarchical decomposition. The intuition behind the interleaved incremental approach is that a relatively small incremental expansion is adequately informed by the preceding decisions, while at the same time avoids forcing extensive commitments to future decisions. Consistent with this intuition, we observed that both very large and very small increments tend to produce solutions of lower quality. In CADET, a typical increment amounts to about 10-20 activities, on the order of 5% of the number of activities in a completed plan.

The incremental expansion process is controlled by a mechanism that leads the algorithm to focus on the most significant and most constrained tasks first, Fig. 8c, and to limit the decomposition to a limited incremental set of tasks. The scheduling step performs temporal constraint propagation (both lateral and vertical within the hierarchy, a partially domain-knowledge driven process), as in Figure 8d, and schedules the newly added activities to the available resources and time periods, shown in Figure 8e.

The temporal constraint propagation component of the scheduling sub-process performs an analysis of all temporal constraints, and adjusts the possible scheduling times of each constrained activity. The temporal representation of an activity consists of an interval of possible start times, an interval of possible end times, and a minimum and maximum duration. Each temporal constraint establishes a relationship between the start or end time of two activities. For example the constraint "SEIZE starts [00:30, 02:00] before SECURE starts," establishes the requirement that the scheduled start time of the SEIZE activity must occur 30 minutes to 2 hours earlier than the scheduled start time of the SECURE activity. The propagation of a temporal constraint sets the possible start or end interval of a constrained activity to the intersection of the initial possible interval with the interval produced when applying the temporal constraint to the other activity in the relationship.

The interleaving approach also meets the users desire to see how the process unfolded.  It allows the user, if desired, to control the number of task expansions. Users could step through the entire process in bite-size expansions consisting of about 10 sub-tasks at a time, and could intervene with interim modifications. However, the significance of this mode of user operation should not be overestimated. Indeed, typically users would step through the process only a few times. After they became familiar with the system and confident in its conclusions, they would thereafter choose to fully expand the plan without performing the step-wise expansion.

With respect to the scheduling sub-process, although we originally planned to use a Multi-decompositional version of the Constrained Heuristic Search [7, 12, 13, 28], we were led eventually to prefer a much simpler and computationally inexpensive scheduling heuristics. These combine domain-independent estimate of the degree to which an activity is constrained, the "earliest-first" rule, and the domain-specific ranking of activity priorities. This choice was driven partly by the rigorous performance requirements, and partly by the fact that the simpler approach tended to produce results more understandable to the users. Generally, given the compound complexity imposed by the need for tight interleaving of multiple, diverse problem-solving processes in CADET, it is prudent to avoid any unnecessary complexity within each of these individual processes.

The same interleaving mechanism is also used to integrate incremental steps of routing, attrition and consumption estimates. A version of the Dijkstra routing algorithm [16] is used to search for suitable routes over the terrain represented efficiently as a parameterized network of trafficable terrain. Optimization can be specified with respect to a number of factors, such as the overall speed of movement or cover and concealment. For estimates of attrition, we developed a special version of the Dupuy algorithm [14] calibrated with respect to estimates of military professionals [29].

The same desire to minimize the complexity of the combined problem-solving process also played a role in our decision to use a no-backtracking approach (with a few minor exceptions). However, even more important considerations had to do with the nature of the problem and the role of the user. We found a user would often find a way to mitigate or to permit infeasibilities (among which the temporal constraint infeasibilities are by far the most numerous; others include insufficient supplies, or unfavorable strength ratios in engagements) by means that are unique to the situation or rely on insights and highly detailed knowledge that go well beyond the limits of CADET's knowledge base.

Furthermore, such resolutions of infeasibilities by the user are usually best done when the user can see the "whole picture" of the plan. For example, reviewing a constraint violation, the user might say, "I am willing to accept this risk in this particular situation in order to achieve my objective." Perhaps it should not be surprising that in a military conflict where both opponents are applying every effort to win, the problem solution is often found at constraint boundaries. We also found that infeasibilities often point to a need for modifications in the original COA, i.e., the sketch-and-statement level of the problem statement, which is human-generated and outside of CADET's purview. All the above considerations led us to dismiss a backtracking-based approach would be largely useless. Instead, CADET continues the solution process even when infeasibility is found, generates the complete (albeit infeasible) solution, displays the solution to the user with infeasibilities highlighted, and then lets the user make the appropriate changes.

### Planning Against a Dynamic Enemy

The adversarial aspects of planning-scheduling problems are addressed via the same incremental expansion mechanism. CADET accounts for adversarial activity in several ways. First, it allows the commander and staff to specify the likely actions of the enemy. The automated planning then proceeds, taking into account, in parallel, both the friendly and enemy actions. Further, CADET automatically infers (using its knowledge base and using the same expansion technique used for hierarchical task network planning) possible reactions and counteractions, and provides for resources and timing necessary to incorporate them into the overall plan. We adopted the Action-Reaction-Counteraction (ARC) heuristic technique used in the traditional COA analysis phase of the Military Decision Making Process [2]. In the Action-Reaction-Counteraction (ARC) approach, an action possible by either friendly or enemy warrants examination for potential reactions. This is followed with further analysis to determine if there exists a counter-action that can be used to minimize the impact of the reaction or negate its effects completely. The ARC technique was augmented with parallel planning for both friendly and enemy forces.

Consider the example of the activity called "forward passage of lines," see Figures 9a and 9b, in which a unit of force passes through the lines of defense manned by a friendly unit and then engages the enemy. When performing this activity, both the unit being passed, and the passing unit, are susceptible to enemy artillery fire. Therefore, CADET's knowledge base includes a method that, in the process of expanding this activity (Fig. 8b), postulates that if the enemy has suitable forces, it will react by attacking by fire the passing unit. The method searches for enemy artillery units within the range from the location of the passage of lines, and creates the (hypothetic) enemy reaction activity "artillery fire" performed by the available enemy unit. This in turn triggers the generation of counteraction activities.

Counteraction activities may be viewed as proactive measures to mitigate risks. The friendly force commander, anticipating an effort by the enemy to concentrate artillery fires during the passage of lines, preplans the appropriate counter-battery fires and has them ready to execute on short notice, enabling him to respond vigorously within seconds of the enemy reaction. To achieve this effect in CADET, the "artillery fire" activity includes a method for generating counteractions, such as a military intelligence activity of locating the enemy artillery that is performing the artillery fire, and a "counter-battery fire" activity against the attacking artillery unit. When the "artillery fire" activity undergoes the expansion process, this method searches for suitable units and, if available, creates and assigns the counteraction activities to them. At the allocation step (Fig. 8d), these reaction and counteraction activities execute their respective methods to estimate the attrition and consumption resulting from the execution of the activities.

Note that the reaction definition is not limited to a single possible reaction as described in the above example. It is possible to define in the knowledge base a set of possible reactions one or several of which are selected based on the state of the battlefield.

ACR does not involve an explicit search, in the sense that it does not explicitly explore multiple alternative moves at each decision point, and it does not involve backtracking, except for the user-driven backtracking as discussed in section "Integration of Multiple Problem Aspects." Instead, the process proceeds in a linear, depth-first fashion (Fig. 9a and 9b): for every newly generated action, the ARC method of the activity (if one is specified within the KB) produces an activity (or activities) representing the enemy reaction; the reaction activity in turn trigger a similar generation of counteraction activities.

To prevent this process from continuing indefinitely, the algorithm terminates at the (heuristically) fixed depth of 3, i.e., an action is followed by reaction which is followed by counteraction, but at this point the chain ends and counteraction is not followed by yet another counter-counter-action.

Justification for this termination heuristic if found in the experience of domain experts, as well as in our experimental observation that going beyond the depth of 3 produces excessive level of details without contributing to the value of the solution for the user. In effect, the cumulative level of uncertainty (will the reaction actually happen when the battle unfolds? Will the counteraction be actually applicable?) becomes so high that planning at a further level of detail is unproductive. Even at the seemingly short cutoff depth of 3, the solutions are often complex and non-obvious to the user.

The ARC process does not search for an optimal solution and does not guarantee one. Rather, the intent is to produce a solution that is consistent with user's expectations and doctrinal training and is produced much faster and more accurately than in the manual process.

Clearly, a suboptimal solution can occur in the ARC algorithm. For example, a reaction produced by the ARC method may turn out to be highly vulnerable, under a specific set of circumstances, to a counteraction. In such a case, this reaction would unlikely to be employed by an intelligent enemy and a large portion of the plan becomes erroneous. .

Although ARC does not guarantee optimality, it produces solutions of the quality that experts find comparable to those of human experts (section "Experimental Evaluation"). The rules that generate reactions and counteractions embody expert knowledge. The probability of generating a grossly suboptimal solution is minimized because the rules that generate a reaction do implicitly account for probable counteractions.

It should be emphasized that the problem addressed by ARC is not directly comparable to a game formulation. In particular, action and reactions are not comparable to moves in a sequential game because they do not need to occur in a sequential order, e.g. a reaction may in fact be executed prior to action, and counteraction – prior to reaction (see Fig. 9a). Nevertheless, the game-theoretic analogy is appropriate here in the sense that an ideal planner in CADET would have to explore the entire game tree (with appropriate pruning applied) in order to be assured of the optimality of a particular action (compare [26]). The ARC approach, however, trades optimality for speed and transparency of solution.

## Knowledge Representation

In the object-oriented fashion, the CADET Knowledge Base (KB) is a hierarchy of classes of activities, see Fig. 13. A class of activities (e.g., Fig. 10) contains a number of procedures (methods) responsible for: computing conditions of applicability of an expansion method; generating sub-activities of an activity depending on such factors as the available assets, the terrain or the location and type of the enemy forces (e.g., Fig 10c); adding temporal constraints; estimating timing and resources required for the activity(e.g., Fig. 10d, 10e, 10g, and 10h); finding suitable routes and locations (e.g., Fig 10f); etc. When there are multiple rules within a procedure (e.g., see Fig. 10c), they are explicitly ordered.

In practice, the most expensive (in terms of development and maintenance costs) part of the Knowledge Base (KB) are the rules responsible for expansion of activities. We find a great variability in the procedures used to evaluate preconditions of expansion and the expansion itself. While some are very simple, others are complex. Many, for example, refer to qualitative geographic and temporal relations between units of force and features of the battlefield and require significant computations using general-purpose programming operators.

Consider (Fig. 11) an example of a complex expansion procedure that applies to a common class of activities in which a military units advances on the battlefield along an axis of advance or a route. When performing an advance, the sub-activities or auxiliary activities of the unit are largely defined by the battlefield state along the route being traversed. For example, the advancing units may encounter enemy units, or obstacles, or require a resupply.

The procedure described in Fig. 11 generates an ordered set of activities based on a projected (estimated for future time points) state of the battlefield. The generation of each activity updates the projected state for use in the generation of future activities. Although the procedure includes elements of sequencing and time-dependent effects, the actual scheduling of the activities generated is delayed until the execution of the scheduling algorithm (see Fig. 8c and Fig. 8d).

The expansion procedure is parameterized to allow the generation of a variety of actions based upon the requirements of the specific activity modeled. For example, the reaction to encountering an enemy unit is different when the advance is performed for the purpose of seizing an objective, as compared to an advance performed for the purposes of a feint attack.

A class of activity has only one expansion procedure, e.g., Fig. 10c, even though a number of different sets of expanded activities may result from an application of a given expansion procedure, depending on the circumstances. For example, the procedure described in Fig. 10c produces different expansions depending of whether the unit is within the suitable distance from the desired minefield location or needs to move toward the location, and whether the unit has enough mines on hand or needs to replenish its stocks.

A different approach would be to encode such alternative expansions as independent rules and let a search process to determine the applicable rules depending on preconditions and on downstream outcomes. We found it preferable to handle all alternative expansions within a single procedure. The number of expansions in this domain tends to be modest and can be more easily and explicitly controlled within a single procedure. This is also consistent with our desire to avoid search and backtracking, as discussed in detail in section "Integration of Multiple Problem Aspects."

CADET includes a module for Knowledge Base (KB) maintenance that allows a non-programmer to add new units of knowledge or over-write old ones, in a relatively simple point-and-click fashion. There is a clear need to allow the end-users an ability to modify the KB in field environments, and the assumption is that the user is not a programmer, not even to the extent of a entering a spreadsheet-like function. Although necessarily limited by our decision to eliminate any direct programming features, the KB maintenance tool does allow an end-user to enter potentially a majority of the required activity classes, albeit clearly not all.

*User Interfaces*

Only a minimal set of necessary user interface features were developed. These consisted mainly of an interface patterned after the synchronization matrix (Fig. 6, 7), allowing the user to click on any cell (activity) and browse through the related network of domain-relevant objects (e.g., the units performing the activity, the location of the unit, etc.). The reason for a minimal interface is that we do not expect CADET to be deployed with a stand-alone user interface, but rather as a part of a larger framework with an existing interface.

Even more important was our realization that the presentation paradigms accepted in the practice of military decision making are not necessarily a good basis for user interfaces in a tool like CADET. In a series of experiments (to be discussed shortly) we found, in particular, that although the synchronization matrix is an accepted way of recording the results of COA analysis, the users of CADET had some difficulty comprehending the synchronization matrix generated by the computer tool, even though it was presented in a very conventional, presumably familiar manner. One possible explanation is that the synchronization matrix works well only as a mechanism for short-hand recording of one's own mental process. However, the same synchronization matrix is not nearly as effective when used to present the results of someone else's, e.g., CADET's, reasoning. The problem was further exacerbated by the fact that CADET-generated matrices were unusually detailed and therefore large, making it difficult for the users to navigate within this large volume of information.

**Experimental Evaluation**

How does planning with CADET compare with purely manual planning? We desired to know what difference the tool would make when put in the hands of the military staff planner.

In the course of interviewing experienced military planners for the purpose of knowledge acquisition, we had been offered varying opinions concerning the potential usefulness of such a tool in the hands of a staff planner. Some described tactical planning as more art than science and dismissed any attempt to automate portions of the process. Some thought the process would greatly benefit planners with minimal experience in the field. Still others, questioned the degree to which the tool would avoid generating predictable (to the enemy) products. They argued that any plan produced in part or in whole by a computerized system would inevitably involve a simplistic and predictable approach. These concerns were addressed via an experiment.

The experiment involved five different scenarios of Brigade-sized offensive operations, nine judges (all Army or Marine field grade officers (Colonel, Lieutenant Colonel or Major), mostly active duty), four types of planning products for each scenario, and three individual grades that the judges were asked to assign to the products.

The brigade is commonly the lowest level with a dedicated planning cell within the headquarters staff. At the same time, even at a relatively low echelon, the military is relying on improvements in decision support tools to improve tactical decision making without increasing the size of battlefield staffs. [15]

The five scenarios were obtained from several exercises conducted by the US Army in recent years. All scenarios differed significantly in terrain, mix of friendly forces, nature of opposing forces, and scheme of maneuver. For each scenario/COA we located the original COA sketches assigned to each planning staff, and the original hand-written synchronization matrices produced by each planning staff. In a few cases, the original COA statement was unavailable and had to be reconstructed based on the sketch and the synchronization matrix.

Although we did not have the specific records of how many officers were involved in these planning sessions and how many hours the sessions lasted, the participants and observers of similar exercises estimated that typically these are performed by a team of 4-5 officers, over the period of 3-4 hours, amounting to a total of about 16 person-hours per planning product.

Taking the five synchronization matrices produced by the human staff, we transcribed them from the original handwritten spreadsheets into a computerized spreadsheet, in order to make them visually less distinct from the computer-generated ones. This set of 5 matrices is referred to as Hand-Human (HH), meaning, "produced by hand and retaining the original human-made style."

The same matrices were then edited to make them look machine-generated. The rationale for doing this was to minimize the possibility of biases of the evaluators against or in favor of computerization. This "disguise" involved standardization of terms and abbreviations, uniform style, elimination of spelling errors, etc. However, all essential content remained unaltered. This set of 5 matrices is referred to as Hand-Machine (HM), meaning "produced by hand and disguised with a machine-looking style."

Using the same scenarios and COAs, we used CADET to generate a detailed plan and a synchronization matrix. The matrices were then reviewed and edited by a surrogate user, a retired US Army officer. The editing was rather light – in all cases it involved changing or deleting no more than 2-3% of entries on the matrix.  This reflected the expected mode of operation – CADET is not expected to be used automatically, but rather in collaboration with a human decision-maker.

CADET was used "as-is." No attempt was made to customize, tailor or extend CADET's knowledge base for these scenarios.

The time to generate these products involved less than 2 minutes of CADET execution, and about 20 minutes of review and post-editing, for a total of about 0.4 person-hours per product. (Although it also took a certain amount of time to enter the sketch-and-statement information into CADET, we do not consider this time because in a realistic digital environment the sketch-and-statement information would be transmitted to the planning staff in a digital format.)

The resulting matrices were transferred to a computerized spreadsheet and given the same visual style as that of Hand-Human (HH) and Hand-Machine (HM) sets. This set is referred to as Computer-Machine (CM), meaning "produced with computer support and retaining the original machine-made style."

The same Computer-Machine (CM) matrices were then edited to give them a "human-made" look but without changing any content. In particular, some common spelling errors were introduced, some "human" inconsistencies in terminology and abbreviations, etc. The rationale, again, was to elicit in our experiments any differences in evaluations that might result from computer-aversion. This set is referred to as Computer-Human (CH), meaning "produced with computer support and disguised with human-made style."

The nine judges, evaluators of the products, were recruited primarily from active duty officers of US Army and Marines, mainly of Colonel and Lieutenant-Colonel ranks, and two recently retired senior officers with extensive experience with current military doctrine. A total of 20 evaluation packages were prepared for the judges: 5 HH, 5 HM, 5 CM and 5 CH. Each package consisted of a sketch, statement, synchronization matrix and a questionnaire that instructed the judges on how to grade the products.

To avoid evaluation biases, assignments of packages to judges were fully randomized. Each judge was asked to evaluate 4 packages. Judges were not told by what means the products were created.

Furthermore, to avoid any biases that could arise from a direct comparison of products, a judge never received any two packages related to the same scenario.

Each judge was asked to review a package and to provide 3 grades: one grade that characterizes the correctness and feasibility of the plan as reflected in the synchronization matrix, on the scale of 1-10; one grade that characterizes the completeness and thoroughness of the plan as reflected in the synchronization matrix, on the scale of 1-10; and one qualitative grade that compares the plan with typical products they see in today's Army practice (ranging from "much worse" to "much better"). Instructions to the judges were worded in a way that associated the grade of 5 with the "typical quality found in today's practice."

Data showed a significant scatter (Fig.12). While mean values for the 4 series ranged from 3.9-5.0, standard deviations ranged from 1.6-2.4. Judges comments also demonstrated significant differences of opinion regarding the same product. We did not find any bias against computerized-looking products.

Some judges strongly disliked the larger and more detailed synchronization matrices produced by CADET. From some judges' comments, it appeared that they assumed that these highly detailed products were generated in a conventional process, and that staff officers were wasting their time and attention by going into such excessive and unusual level of detail. Perhaps this suggests that merely by reducing the size and level of details shown on the synchronization matrix (which can be easily done automatically), one could improve the grades given to the CADET's products.

Overall, the results (Fig.12) demonstrated that CADET performed on par with the human staff - the difference between CADET's and human performance was statistically insignificant. Taking the mean of grades for all five scenarios, CADET earned 4.2, and humans earned 4.4, with standard deviation of about 2.0, a very insignificant difference. Finally, comparing the "undisguised" results, i.e., series CM and HH, CADET earned the mean grade of 4.4 and humans earned 3.9, although the difference is still rather small. The conclusion: CADET helps produce complex planning products dramatically (almost two orders of magnitude) faster yet without loss of quality, as compared to the conventional, manual process.

**Experiments in a More Realistic Setting**

Another experimental exploration [3] placed the CADET component in a more realistic setting, integrated in a suite of decision aids and suitable for performing a complete cycle of the Army Brigade decision-making. Sponsored and performed by an Army Battle Command Battle Lab, this set of experiments approximated the use of battle planning aids by a group of officers in a Brigade Command Post.

The experimenters focused on broader issues associated with potential negative impacts of such tools on the creative aspects of the art of war. Will such tools encourage excessive fixation on analytical aspects of command, by the book and by numbers? Will they detract from intuitive, adaptive, art-like aspects of the military command decision-making? Will it make the plans and actions more predictable to the enemy?

The experiments involved two teams, each consisting of eight Army officers (majors and lieutenant colonels) at Battle Command Battle Laboratory-Leavenworth facilities in Fort Leavenworth, Kansas. All the subjects were from combat arms branches and had a variety of tactical experience, with 11-23 years of Active Service. One team performed a typical process of decision-making using the traditional manual method. The other team accomplished the same tasks with the support of the decision aids, particularly CADET.

The observations during the processes of both teams, and the comparison of the products dispelled the concerns. The quality of the products generated by both teams was comparable even though the computer-aided team completed its tasks much faster. There was no evidence that the decision aids would in any way encourage a cookbook approach. To the contrary, because they allow the planners to explore rapidly a broader range of possible courses of action, including those that are unconventional, there is a potential for such tools to encourage greater ingenuity, creativity and adaptivity.

**Lessons Learned and Future Directions**

CADET is a tool developed to generate Army battle plans with realistic degree of detail and completeness, for multiple battle operating systems, for the large scale and scope associated with such complex organizations as an Army Division or a Brigade. CADET performs dramatically faster than a conventional human-only planning staff, with comparable quality of planning products.

CADET shows a promise of reaching the state where a military decision-maker, a commander or a staff planner, uses it in field conditions, as a part of an integrated suite of tools available on a rugged, highly portable computer device. He would use it routinely to perform planning of tactical operations, to issue operational plans and orders, and to monitor and modify the plans as the operation is executed and the situation evolves.

The CADET technology was designed for a particular application domain – military operations planning. The results of our experiments, as well as the interest in the CADET technology expressed by several organizations within the US military, offer evidence that the technology is likely to be of value in its primary intended  application domain. The military planning domain is in itself a very large market, with a broad range of sub-domains and problems (e.g., pre-operation planning, in-execution management, simulation, training, military robotics, etc.), and with a major amount of funding dedicated annually to introducing and improving intelligent applications, many of which can leverage the CADET technology.

Outside of the military applications world, CADET's technical approach may also be considered for use in other domains that require either a close integration of planning and scheduling and movement routing, or explicit planning for potential adverse effects of actions and the appropriate precautions, i.e., a kind of adversarial reasoning. The former category may include for example construction industry planning, especially in multiple geographically dispersed projects. The latter category may include law enforcement, disaster response (an exploratory prototype using CADET has been already constructed), business campaign planning (market penetration, new product release, marketing campaigns, etc.), and industrial or space robotics (compare with [24] where an adversarial planner was used for a space robotics application.) CADET-like approach may also serve as a component of solution to a much broader problem of society-wide activities, e.g., international interventions, that involve not only military but also long-term political, economic, social and ideological processes [30].

A number of practical lessons have been learnt:

-- Adopting or developing a simple and transparent concept of user operation helps to increase the probability of user acceptance and to reduce the developmental risks and training requirements. Such a concept does not necessarily coincide with (and in fact is likely to be significantly different from) the established ways of performing similar processes in a manual fashion.

-- The problem of battle planning is an instructive example of a real-world problem where distinct computational problems are closely coupled and do not lend themselves to a useful decomposition. One effective, practical approach to such problems can be the tight interleaving of incremental planning, scheduling, routing and attrition and consumption calculations.

-- Given the complexity of interleaving multiple, diverse computational problem-solving processes, it is prudent for developers to reduce technical risks by using less complicated algorithms in each of the individual processes. Computationally inexpensive algorithms that often trade optimality for speed help to assure an almost instantaneous response to the user, which is important in military applications.

-- The Action-Reaction-Counteraction (ARC) heuristic can be a practical, robust technique for integrating adversarial considerations into the planning process.

-- Conventional presentation paradigms rooted in manual procedures are not always a suitable basis for an effective user interface. Rigorous separation – both architectural and conceptual - of problem solving components from user interaction mechanisms allows for integration with a variety of user-interface paradigms.

Key gaps that must be overcome to realize the full potential of such tools:

Military decision-making commonly requires collaboration of multiple officers with distinct functions, responsibilities and expertise. In the near-future warfare, these officers will often collaborate while dispersed over the battlefield, communicating over the tactical internet, possibly asynchronously. Tools like CADET must support such forms of collaboration.

Effective human-machine interfaces remain challenging, especially for complex, multi-dimensional information such as plans and execution of military operations, in high-tempo, high-stress environments, in physically challenging field conditions. Today's common paradigms – map-based visualizations of spatial information and synchronization matrix for temporal visualization – are not necessarily the most promising directions, and different approaches ought to be explored.

Presentation of CADET's products requires qualitatively different user interfaces and visualization mechanisms. Our experiments suggest that the users had difficulties comprehending a computer-generated synchronization matrix, even though it was presented in a very conventional, familiar manner. Perhaps, the synchronization matrix functions well only as a mechanism for short-hand recording of one's own mental process. However, the same synchronization matrix is not nearly as useful when used to present the results of someone else's, e.g., a computer tool's, reasoning.

Combat planners must be able to plan rapidly, communicate orders to subordinates clearly and react without delay to changes in the situation. Planning tools like CADET should give the commander the ability to accelerate this cycle of recognition, re-planning and reaction, i.e., the capability of continuous re-planning during execution.

**Acknowledgements**

The work described in this paper was supported by funding from US Army CECOM (DAAB 07-96-C-D603 - CADET SBIR Phase I, DAAB 07-97-C-D313 – CADET SBIR Phase II and DAAB 07-99-C-K510 - CADET Enhancements); DARPA (DAAB07-99-C-K508 Command Post of the Future); and TRADOC BCBL-H (GS-35-0559J/DABT63-00-F-1247). TRADOC BCBL-L provided additional funding under DAAB 07-99-C-K510.  MAJ R. Rasch of the US Army BCBL-L provided important advice to the CADET team.

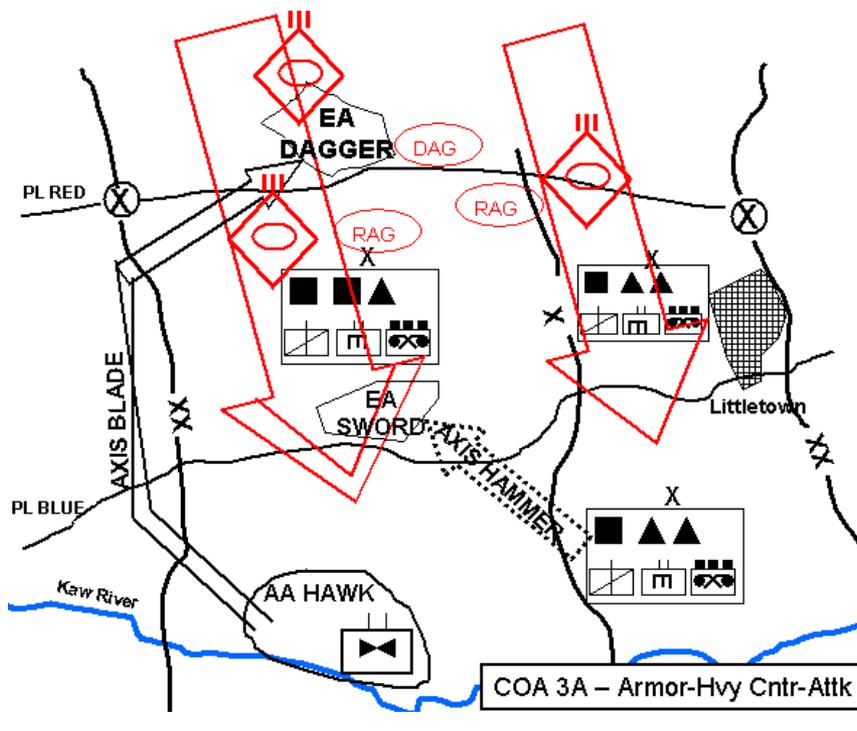

Purpose: Defeat the 2GA attack and prevent penetration of PL BLUE
Decisive Point: Counterattack to defeat the 2TD

|  | |
|---|---|
| ⌀ | **SECURITY**<br>T: CAV SQDRN screens north in sector.<br>P: Delay approach of 1st ech TRs by 6 hours. |
| X<br>■ ▲▲<br>◰ m ⟠<br>**WEST** | **CLOSE/MAIN BATTLE**<br>T: BDE(WEST) Delay and defeat 1st ech TR in sector<br>P: To prevent enemy penetration of PL BLUE<br>T: Delay 2nd ech TR in sector and halt progress forward of PL BLUE.<br>P: Position enemy for counter-attack in EA SWORD |
| X<br>■ ■▲<br>◰ m ⟠<br>**EAST** | T: BDE(EAST) Defeat enemy in sector.<br>P: Cause commitment of 2nd ech in West to facilitate destruction by counter-attack. |
| X<br>■ ▲▲<br>◰ m ⟠<br>**CATK** | T: BDE(CATK) Counter-attack to defeat 2nd ech TR in EA SWORD<br>P: Prevent penetration of PL BLUE and position div for hasty defense |
| ►◄ | **DEEP**<br>T: Move along AXIS BLADE to engage 2nd ech TR in EA DAGGER. Attrite by 35%.<br>P: Slow enemy approach and reduce strength prior to entering MBA. |
| ◉ | T: MLRS fires disrupt 2nd echelon TR movement between EA DAGGER and PL RED<br>P: Prevent enemy from reorganizing and facilitate |

**Figure 1. An example of a (partial) sketch and statement of a course of action.**

| COA 3A | H-0:15 | H-0 | H+0:15 | H+0:30 | H+0:45 | H+1:00 | H+1:45 | H+1:53 | H+1:59 | H+2:03 | H+2:07 |
|---|---|---|---|---|---|---|---|---|---|---|---|
| **THREAT ACTION** | | Enemy main body reaches PL RED | En crosses PL RED | | En commits 2nd ech TR to left AA | En 2d ech TR approaches EA DAGGER | En crosses DP 17 | En nears EA DAGGER | En enters EA DAGGER | En completely within EA DAGGER | En attrited 30% |
| **DECISION POINTS** | | 2, 8 | 4, 7 | | | | 17 | | 18 | 17 | |
| **INTEL** | | | | | | | | | | | |
| **FOCUS** | UAV 1 focus on passage pts | Determine enemy Main Effort | Id enemy commitment of 2nd ech | Launch UAV 4 & 5 | Divert UAV 2 & 3 to overlap coverage of DP 17 | Determine speed, direction of 2nd ech TR | Overlapping UAV coverage of DP 17 | Verify speed, direction | Determine depth of en formation | Locate rear of en formation | |
| **NAI** | 23, 55 | 24, 56 | | | | 17, 43 | 17, 43 | 17, 18, 43, 44 | 18, 44 | 17, 43 | |
| **TAI** | 11, 12 | 14, 15 | | | | 2, 8 | 2, 8 | | | | |
| **MANEUVER** | | | | | | | | | | | |
| **SECURITY** | Cav Sqdrn performs rearward passage of lines along PL RED. | Cav Sqdrn moves to screen left flank from PL BLUE to PL BLACK | | Cav Sqdrn in position | | | | Left BDE fires mortar and DS arty SEAD along left side | | | |
| **MAIN EFFORT** | | Left BDE defends in sector. | | Left BDE falls back, drawing 1st ech TR forward | Left BDE slows rearward movement, stabilize def fwd of PL BLUE | | | Defense along MBA continues | | | |
| **SUPP. EFFORT** | | Right BDE defends in sector. | | Right BDE defends as far forward as possible | Maintain contact with left BDE to avoid pen along bndry | | | Maintain contact to avoid en shifting forces to west | | | |
| **RESERVE** | | A/3-21 ARM in AA SMITH | | BIP to reinforce right BDE to show strength in East | BIP to reinforce left BDE to avoid pen PL BLUE | | | | | | |
| **REAR** | | Accept risk with no TCF | | | | | | | | | |
| **DEEP** | | Pri: delay 2nd ech TRs | | | | | Prep ATK HEL BN for deep attack | Launch ATK HEL BN | ATK HEL crosses FLOT | ATK HEL arrive in defilade psn EA DAGGER | CAS missions strike EA DAGGER | ATK HEL attks 2nd ech TR in EA DAGGER |
| **FIRE SUPPORT** | Pri: suppress AT during POL | Pri: right BDE. Demonstrate stronger def in East | | | | Pri: left BDE. | Prep SEAD | SEAD | MLRS fires in EA DAGGER | | MLRS fires target AA 1C, 1D |

**Figure 2. An example synchronization matrix (partial) produced in a conventional, manual process, starting with the sketch and statement of Fig.1. Such products are usually drawn by hand on a preprinted template of a synchronization matrix. More recently, these are commonly produced with a personal computer, using conventional programs for office presentation graphics and spreadsheets.**

**Figure 3. A COA sketch developed in one of several COA-editing tools that have been used as data-entry interfaces to CADET. A sketch typically includes symbols of friendly and enemy units, assembly areas, objectives, mobility corridors, axes of attacks, etc.**

**Figure 4. An example of the input to CADET, a representation of a traditional sketch and statement.**

Units of force:
        Unit('Blue-Bn-1', Blue, Infantry, Battalion, initalLocationCoordinates = ..., strength = 1.00)
        Unit('Red-Co-1', Red, Armor, Company, initialLocationCoordinates = ..., strength = 0.80)
        ... *(typically a total of 5-50 units)*
Control measures:
        ControlMeasure('Sword', type = ObjectiveArea, boundaryPoints = {...})
        ControlMeasure('mc-1', type=Mobility Corridor, trafficability = high, segmentPoints = {...})
        ... *(typically 50-500 entities)*
Activities:
        Activity('Seize-1', type=seizeObjective, Blue, candidateUnits = {Blue-Bn-1, Blue-Bn-4... },
        objectiveArea = Sword)
        Activity('Attack-1', type=attack, Blue, targets = {Red-Co-1}, intent = attrit)
        ... *(typically 2-20 activities, including both blue and red activities)*
 Temporal constraints:
        Constraint(Seize-1, ends, range = {0min, 2hours}, before, Attack-1, starts)
        ... *(typically 2-20 constraints)*

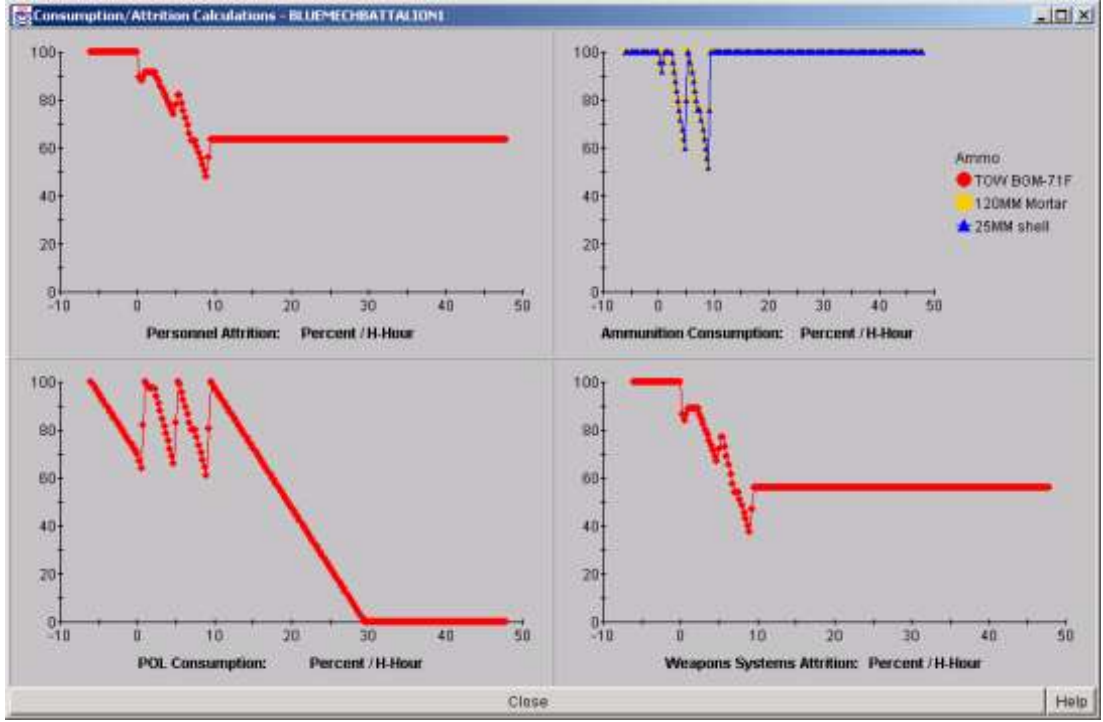

**Figure 5a.  Planners face the need to perform detailed logistical analysis including projected ordnance and fuel consumption and supply chain planning at several echelons. These graphs show the results of the logistical analysis, and attrition estimation for a single unit over the course of the plan.**

| | | | | | | | |
|---|---|---|---|---|---|---|---|
| **POL** | 87% | 86% | 86% | 85% | 82% | 79% | 76% | 73% |
| **TOW BGM-71F** | 96% | 96% | 96% | 96% | 92% | 87% | 83% | 79% |
| **120MM Mortar** | 96% | 96% | 96% | 96% | 92% | 87% | 83% | 79% |
| **25MM shell** | 96% | 96% | 96% | 96% | 92% | 87% | 83% | 79% |
| **1-42 Consumption/Attrition Calculations** | | | | | | | | |
| **Weapons System** | 92% | 92% | 92% | 92% | 92% | 88% | 83% | 78% |
| **Personnel** | 94% | 94% | 94% | 94% | 94% | 90% | 87% | 83% |
| **POL** | 99% | 98% | 98% | 98% | 97% | 94% | 91% | 88% |
| **120MM HEAT-MP-T M830** | 100% | 100% | 100% | 100% | 100% | 96% | 92% | 87% |
| **SABOT: 120MM APFSDS-T M829A1** | 100% | 100% | 100% | 100% | 100% | 96% | 92% | 87% |
| **120MM Mortar** | 100% | 100% | 100% | 100% | 100% | 96% | 92% | 87% |
| **1-43 Consumption/Attrition Calculations** | | | | | | | | |
| **Weapons System** | 87% | 85% | 84% | 83% | 81% | 80% | 79% | 77% |
| **Personnel** | 90% | 89% | 88% | 87% | 86% | 85% | 84% | 83% |
| **POL** | 64% | 61% | 58% | 55% | 52% | 49% | 46% | 42% |
| **120MM HEAT-MP-T M830** | 59% | 55% | 51% | 47% | 43% | 39% | 35% | 31% |
| **SABOT: 120MM APFSDS-T M829A1** | 59% | 55% | 51% | 47% | 43% | 39% | 35% | 31% |
| **120MM Mortar** | 59% | 55% | 51% | 47% | 43% | 39% | 35% | 31% |
| **RES Consumption/Attrition Calculations** | | | | | | | | |
| **Weapons System** | 100% | 100% | 100% | 100% | 100% | 100% | 100% | 100% |
| **Personnel** | 100% | 100% | 100% | 100% | 100% | 100% | 100% | 100% |
| **POL** | 80% | 78% | 77% | 76% | 75% | 73% | 72% | 71% |

**Figure 5b. In this display of the logistics analysis and attrition estimates for all resources together, the human planner's attention is drawn to the shortage of ammunition (red and black cells). Without human intervention in the planning process, this unit will become critically low on ammunition over time.**

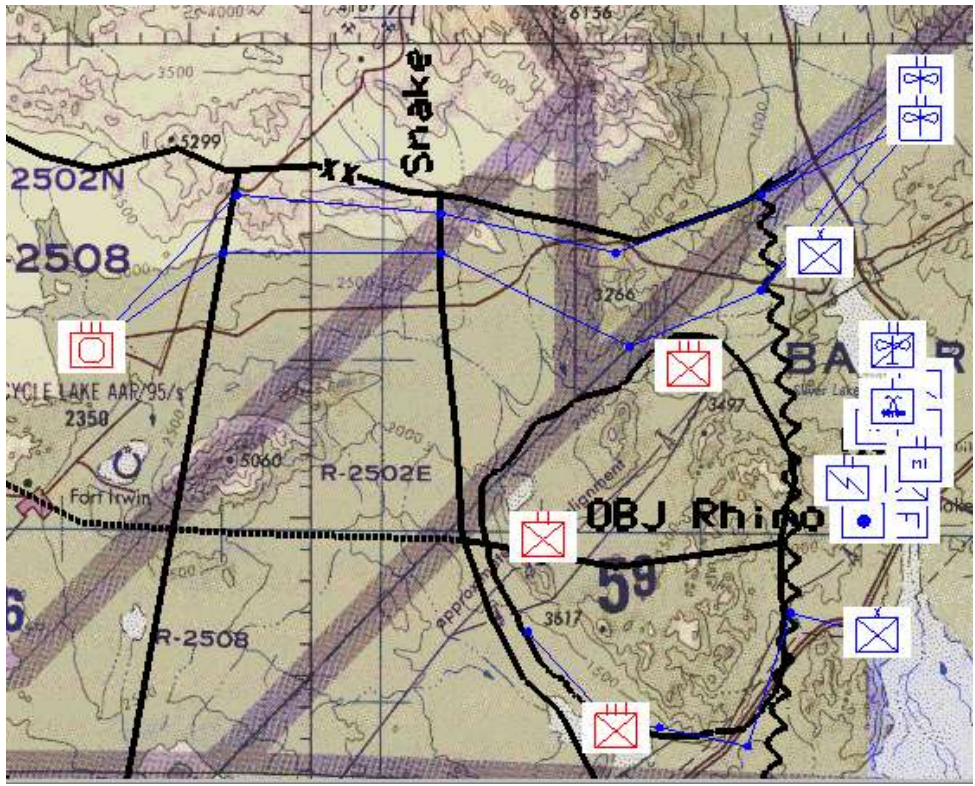

**Figure 6. CADET takes a sketch and statement as an input, and produces detailed plan/schedule of hundreds of tasks, usage of resources, risks and losses, actions of the enemy, and routing.**

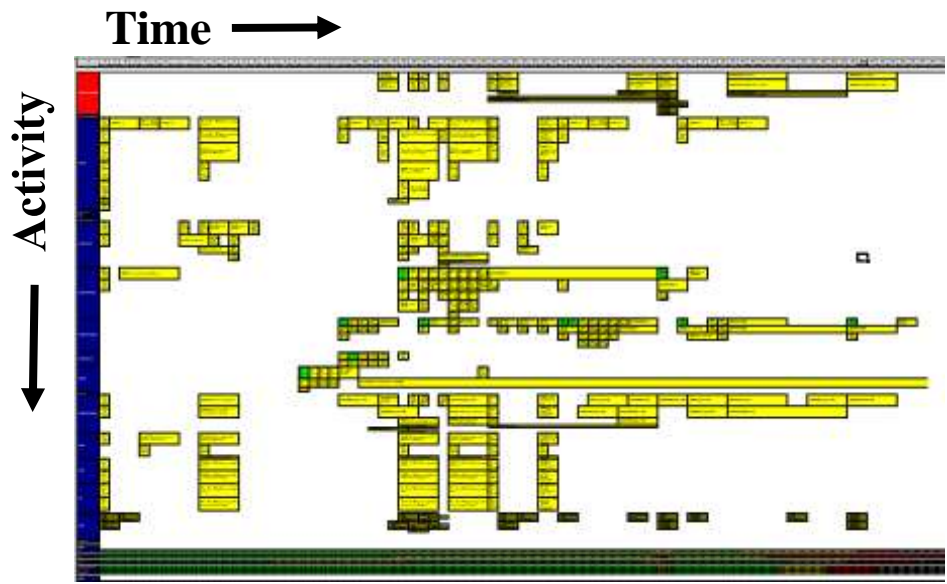

**Figure 7. The main output of CADET is a large synchronization table (a small fraction of a typical product is shown here) that shows a function a of time the recommended tasks of the friendly force, projected activities of the enemy, and status of resources for the participating units of the force.**

**Figure 8a. The algorithms used to produce a plan, and to produce a partial plan.**

**procedure** ProducePlan()
1.　　　**while** planning is not complete **do**
2.　　　　a*ctList* ← list of unexpanded, unallocated activities
3.　　　　ProducePartialPlan(*actList*, 10)

**procedure** ProducePartialPlan(**Activities** *A,* **Integer** *maxCount*)
1.　　　*newActivityCount* ← 0
2.　　　**while** $A \neq \varnothing$ **and** *newActivityCount* < *maxCount* **do**
3.　　　　*highestPriorityAct* ← GetHighestPriorityAct(*A*)
4.　　　　**if** *highestPriorityAct* requires expansion **then**
5.　　　　　*newActs* ← Expand(*highestPriorityAct*)
6.　　　　　*newActivityCount* ← │ *newActs* │ + *newActivityCount*
7.　　　　　*A* ← *newActs* ∪ *A*
8.　　　　PropagateTimeConstraints(*highestPriorityAct*)
9.　　　　**foreach** Activity *act* ∈ *newActs* **do**
10.　　　　　PropagateTimeConstraints(*act*)
11.　　　**while** *highestPriorityAct* is not ready for scheduling **do**
12.　　　　*highestPriorityAct* ← GetHighestPriorityAct (*A*)
13.　　　Schedule(*highestPriorityAct*)

**Figure 8b. The algorithm describing the general steps of an activity expansion. The value returned is the set of all activities generated during the expansion.**

**function** Expand(**Activity** *expandingAct*, **Time** *earliestStartTime*, **BattlefieldState** *S*) : **Activities**
1.     $S'$← a projection of the battlefield state *S* at *earliestStartTime*.
2.     *candidateResources* ← CalculateCandidateResources($S'$.AllResources)
3.     *startPt* ← CalculateStartPoint(*candidateResources*, *earliestStartTime*, $S'$)
4.     *endPt* ← CalculateEndPoint($S'$)
5.     **if** *startPt* != *endPt* **then**
6.       *aPath* ← CalculatePath(*startPt*, *endPt*, $S'$.DirectedGraph, *earliestStartTime*)
7.       *advanceActs* ← Advance(*candidateResources*, *aPath*, *earliestStartTime*, $S'$)
8.       update $S'$to reflect the projected effects of *advanceActs*
9.     *newActs* ← additional activities, reaction, and counteraction derived from *expandingAct*
10.   create time constraints between *expandingAct, newActs*, and *advanceActs*
11.   **return** *newActs* ∪ *advanceActs*

**Figure 8c. Retrieve the highest priority activity.**

**function** GetHighestPriorityAct(**Activities** *A*) : **Activity**
1.       *highestPriorityAct* ← null
2.      **foreach Activity** *act* ∈ *A* **do**
3.         **if** *act* is more temporally restricted than *highestPriorityAct*
4.           **or** CalculatePriority(*act*) > CalculatePriority(*highestPriorityAct*)
5.           **or** [CalculatePriority(*act*) = CalculatePriority(*highestPriorityAct*)
6.           **and** *act*.LatestEndTime > *highestPriorityAct*.LatestEndTime ]
7.         *highestPriorityAct* ← *act*
8.      **return** *highestPriorityAct*

**Figure 8d. Propagate temporal constraints for an activity.**

**procedure** PropagateTimeConstraints(**Activity** *fromAct*)
1.      **foreach Constraint** *tc* ∈ *fromAct*.AllConstraints **do**
2.          *toAct* ← *tc*.getOtherActivity(*fromAct*)
3.          **if** *tc* has not been propagated in this propagation cycle **and** *toAct* is not scheduled **then**
4.              *ci* ← the timing interval, start or end, of *fromAct* that is constrained by *tc*
5.              *si* ← the interval of possible start times for *toAct*
6.              *ei* ← the interval of possible end times for *toAct*
7.              **if** *tc* effects *si* **and** *tc* applied to any value in *ci* produces a value outside the interval *si* **then**
8.                  *si* becomes the intersection of *si* and the interval produced from the application of *tc* over *ci*
9.              **if** *tc* effects *ei* **and** *tc* is not valid for *ei* and *ci* **then**
10.                 *ei* becomes the intersection of *ei* and the interval produced from the application of *tc* over *ci*
11.             PropagateTimeConstraints(*toAct*)

**Figure 8e. The algorithm used to schedule an activity.**

**procedure** Schedule(**Resources** *candidateResources*, **Activity** *actToSchedule*)
1.      **while** *actToSchedule* is not scheduled **and** *candidateResources* list is not empty **do**
2.       *duration* ← CalculateDuration(*candidateResources$_0$*, *actToSchedule*)
3.      **for Time** *possibleStartTime* **from** *actToSchedule*.EarliestStart **to** *actToSchedule*.LatestStart **do**
4.         *endTime* ← *possibleStartTime* + *duration*
5.         **if** *endTime* is within [*earliestEndTime*, *latestEndTime*]
6.             **and** *candidateResources$_0$* has no assignments over [*possibleStartTime*, *endTime*] **then**
7.           RecordConsumption(*startTime*, *endTime*, *candidateResources$_0$*)
8.           schedule the activity from *startTime* to *endTime*
9.         **else**
10.         Remove(*candidateResources$_0$*, *candidateResources*)

**Figure 9a. A typical example of the series of expansions which produce an action, reaction, and counteraction (ARC). The activity initially expanded is a passage of lines with an expansion as shown in A. The reaction activity is produced when generating the ARCs of the passage of lines, shown in B, and is temporally constrained to occur prior to the action. The generation of ARCs for the enemy fire on the passage point produces the two counteractions shown in C.**

**Figure 9b. Two examples of action-reaction-counteraction calculation methods. The first method called during the expansion of a passage of lines generates and synchronizes an artillery fire. The second is called during the expansion of the artillery fire, and generates a counteraction consisting of counterbattery activities.**

**function** GenerateActionReactionCounteraction(**Activity** *passageAct,* **BattlefieldState** *S*) : **Activities**
*1.*      *candidateResources* ← CalculateCandidateResources(*S*.AllResources)
2.      **If** *candidateResources* exist **and** *candidateResources* are in range of *passageAct.*Location **then**
3.         *fireOnPP* ← new Activity (ARTILLERY FIRE, *candidateResources*)
4.         CreateTimeConstraint(*fireOnPP*, STARTS, interval=[00:15, 00:30], BEFORE, *passageAct*, STARTS)
5.         **return** { *fireOnPP* }

**function** GenerateActionReactionCounteraction(**Activity** *fireOnPP,* **BattlefieldState** *S*) : **Activities**
1.      *candidateResources* ← CalculateCandidateResources(*S*.AllResources)
2.      **If** *candidateResources* exist **and** *candidateResources* are in range of *fireOnPP.*Location **then**
3.         *counterbattery* ← new Activity (ARTILLERY FIRE, *candidateResources*)
4.         *findEnemy* ← new Activity (FIND ENEMY, *candidateResources*)
5.         CreateTimeConstraint(*counterbattery*, STARTS, interval=[00:00, 00:00], BEFORE, *fireOnPP*, STARTS)
6.         CreateTimeConstraint(*findEnemy*, STARTS, interval=[00:00, 00:00], BEFORE, *fireOnPP*, STARTS)
7.         CreateTimeConstraint(*counterbattery*, ENDS, interval=[00:00, 00:00], BEFORE, *fireOnPP*, ENDS)
8.         **return** { *findEnemy, counterbattery* }

**Figure 10a. An example of an activity description.**

**Activity Class Name**: Emplace a Scatterable Minefield
**Required Inputs:** The location, width, and intent of the minefield.
**Scheduling and allocation priority (Figure 10b):** High priority if the activity is a part of the Main Effort, medium otherwise.
**Expansion Method (Figure 10c):** Create a series of activities that move the unit into a position in firing range of the desired minefield location. If additional ordnance is required generate an activity of supplying the required ordnance,. Create temporal constraints on the sub-activities.
**Method to estimate the duration of this activity (Figure 10d):** Calculate using a combination of tables and algebraic formulae based on the number of vehicles in the unit capable of firing the required ordnance, the intent and dimensions of the minefield, and the rate at which the unit can fire the specified ordnance type.
**Method to assign a resource (or a set of candidate resources) to this activity (Figure 10e):** If available utilize artillery provided by the division. If divisional artillery is unavailable utilize the artillery unit organic to the main effort of the COA.
**Method to determine how much of the resource is required:** Estimate the minimum size or fraction of the unit required to perform this activity based on a combination of tables and algebraic formulae considering the size of the minefield, its intent and type, capabilities of the unit, etc.
**Method to find start point of the activity (Figure 10f):** Find the point where the assigned unit is located after completing the most recent preceding activity.
**Method to find the end point (Figure 10f):** Find the point closest to the start point that is in firing range of the desired minefield location.
**Path finding method:** Use the standard routing (a version of Dijkstra) algorithm with a weighting scheme favoring paths with low threat and high accessibility.
**Ammo consumption estimation method (Figure 10g):** Calculate using a formula involving the desired density, and dimensions of the minefield.
**Attrition estimation method:** Calculate using a formula relating the fraction of personnel and equipment losses to the type of an engagement with hostile forces, the strengths of the forces, posture, environmental conditions, etc.

**Figure 10b. An example priority calculation algorithm as defined for a minefield emplacement activity**

**function** CalculatePriority(**Activity** *emplaceAct,* **Resource** *aResource*) :
        **ActivityPriority**
1.        **if** *aResource* is a divisional resource **then**
2.          **return** high priority
3.        **if** *emplaceAct* is a part of the main effort **then**
4.          **return** medium priority
5.        **else if** *emplaceAct* is a part of the supporting effort
6.          **or** *aResource* is a subordinate unit resource **then**
7.          **return** low priority
8.        **return** medium priority

**Figure 10c. An example expansion algorithm as defined for a minefield emplacement activity.**

**function** Expand(**Activity** *expandingAct*, **Time** *earliestStartTime,* **Points** *minePts*) : **Activities**

1.    $S'\leftarrow$ a projection of the battlefield state $S$ at *earliestStartTime*
2.    *candidateResources* $\leftarrow$ CalculateCandidateResources($S$.AllResources)
3.    *startPt* $\leftarrow$ CalculateStartPoint(*candidateResources*, *earliestStartTime*, $S'$)
4.    *endPt* $\leftarrow$ CalculateEndPoint(*startPt*, *minePts*, $S'$.DirectedGraph, *candidateResources*)
5.    **if** *startPt* $\neq$ *endPt* **then**
6.        *moveAct* $\leftarrow$ new Activity (MOVE, *candidateResources, startPt, endPt*)
7.        CreateTimeConstraint(*moveAct*, ENDS, interval=[00:00, ∞], BEFORE, *expandingAct*, STARTS)
8.        *newActivities* $\leftarrow$ *newActivities* $\cup$ { *moveAct* }
9.        update $S'$to reflect the effects of *moveAct*
10.    **if** ordnance levels are below required levels **and** fuel is below required level **then**
11.        *supplyAct* $\leftarrow$ new Activity(FULL RESUPPLY, *candidateResources*)
12.    **else if** fuel is below required level
13.        *supplyAct* $\leftarrow$ new Activity(BASIC REFUEL, *candidateResources*)
14.    CreateTimeConstraint(*supplyAct*, ENDS, interval=[00:00, 00:30], BEFORE, *moveAct*, STARTS)
15.    *newActivities* $\leftarrow$ *newActivities* $\cup$ { *supplyAct* }
16.    **return** *newActivities*

**Figure 10d. An example duration calculation algorithm as defined for a minefield emplacement activity.**

**function** CalculateDuration(**Resource** *aResource*, **MinefieldIntent** *mIntent*, **Number** *mWidth*, **MinefieldType** *mType*) : **Duration**
1.    *numAPts* ← GetNumberOfAimingPoints(*mType*, *mWidth*)
2.    *density* ← GetRequiredDensity(*mIntent*)
3.    *rndsPerAPt* ← GetNumberOfRoundsPerAimingPoint(*density*, *mType*)
4.    *numVehicles* ← GetNumberOfVehiclesInUnit(HOWITZER, *aResource*)
5.    *firingRatePerVehicle* ← GetFiringRate(HOWITZER)
6.    **return** (*rndsPerAPt* \* *numAPts*) / (*numVehicles* \* *firingRatePerVehicle*)

**Figure 10e. An example candidate resource calculation algorithm as defined for a minefield emplacement activity**

**function** CalculateCandidateResource(**Resources** *allResources*) : **Resources**
1.      *candidateResources* ← Empty List
2.      **foreach Resource** $r \in$ *allResources* **do**
3.          **if** $r$ is on the side required for the activity **and**
4.              $r$ can perform artillery tasks **and**
5.              $r$ is provided by the division **then**
6.              Insert ($r$, *candidateResources*)
7.          **else if** $r$ is on the side required for the activity **and**
8.              $r$ can perform artillery tasks **and**
9.              $r$ is a subordinate to the main effort of the COA **then**
10.             Append($r$, *candidateResources*)
11.     **return** *candidateResources*

**Figure 10f. Example start and end point calculation algorithms as defined for a minefield emplacement activity**

**function** CalculateStartPoint(**Resource** *aResource*, **Time** *startTime,* **BattlefieldState** *S*) : **Point**
1.　　　　**return** the position of *aResource* at time *startTime*

**function** CalculateEndPoint(**Point** *startPoint*, **Points** *minefieldPoints*, **DirectedGraph** *dGraph*,
　　　　**Resource** *aResource*) : **Point**
1.　　*sortedPointList* ← SortByProximityToPoint(*dGraph*.Points, *startPoint*)
2.　　$i \leftarrow 0$
3.　　**while** *endPoint* is null **do**
4.　　　**if** DistanceBetween (*sortedPointList$_i$*, *minefieldPoints*) < *aResource*.FiringRange
5.　　　　**and** a path exists between *sortedPointList$_i$* and *minefieldPoints* **then**
6.　　　　*endPoint* ← *sortedPointList$_i$*
7.　　　$i$++
8.　　**return** *endPoint*

**Figure 10g. Example fuel and ammunition consumption algorithm as defined for a minefield emplacement activity**

**procedure** RecordConsumption(**Time** *startTime*, **Time** *endTime*, **Resource** *aResource*)
1.       *density* ←GetRequiredDensity(*mIntent*)
2.       *numAPts* ← GetNumberOfAimingPoints(*mType*, *mWidth*)
3.       *totalRequiredRounds* ← GetNumberOfRoundsPerAimingPoint(*density*, *mType*) * *numAPts*
4.       RecordConsumption(*aResource*, *startTime*, *endTime*, MINEFIELD_ORDNANCE, *totalRequiredRounds*)
5.       RecordConsumption(*aResource*, *startTime*, *endTime*, FUEL, NON_MOVING_FUEL_USAGE_RATE)
6.       RecordConsumption(*aResource*, *startTime*, *endTime*, STANDARD_ORDNANCE, BACKGROUND_STD_ORDNANCE_RATE)

**Figure 10h. Example attrition estimation algorithm as defined for an Attack activity. The calculation of the values for personnel and weapon systems attrition in steps 6 through 9 use the empirical coefficients derived partly from Dupuy [14].**

**procedure** RecordAttrition(**Time** *startTime*, **Time** *endTime*, **Resource** *aResource,* **Units** *targets*)
1.  *friendlyAttritionFactor* ← An activity dependant factor applied to the friendly attrition
2.  *defStrength* ← 0
3.  **foreach Unit** *u* ∈*targets* **do**
4.   *defStrength* ← *defStrength* + StrengthAtTime(*u, startTime*)
5.  *attStrength* ← StrengthAtTime(*aResource, startTime*)
6.  *attPersonnelAttrition* ← *K * attPostureFactor*\*[*(defStrength\*terrainFactor\*defPostureFactor)/attStrength*]$^{0.41}$
7.  *defPersonnelAttrition* ← *K * defPostureFactor*\*[*(defStrength\*terrainFactor\*defPostureFactor)/attStrength*] $^{-0.41}$
8.  *attWSAttrition* ← *c * attPersonnelAttrition*
9.  *defWSAttrition* ← *c * defPersonnelAttrition*
10. **foreach Unit** *u* ∈*targets* **do**
11.   RecordAttrition(*u, startTime, endTime,* WEAPONS_SYSTEMS, *defWSAttrition*)
12.   RecordAttrition(*u, startTime, endTime,* PERSONNEL, *defPersonnelAttrition*)
13.  RecordAttrition(*aResource, startTime, endTime,* WEAPONS_SYSTEMS, *attWSAttrition\* friendlyAttritionFactor*)
14.  RecordAttrition(*aResource, startTime, endTime,* PERSONNEL, *attPersonnelAttrition\* friendlyAttritionFactor*)

**Figure 11. Expansion procedure for the activity class Advance. This function returns an ordered list of activities required to accomplish the advance.**

**function** Advance(**Resource** *aResource*, **Path** *aPath*, **Time** *projectedStart*, **BattlefieldState** *S*) : **Activities**

1. $S'\leftarrow$ a projection of the battlefield state $S$ at *projectedStart*
2. *newActs* $\leftarrow$ empty list
3. $n \leftarrow 0$
4. **for** $i$ **from** 0 **to** *aPath*.NumSegments **do**
5.  *requiredActType* $\leftarrow$ GetRequiredActivityType( *aPath*.Segments$_i$, $S'$)
6.  **if** *newActs$_n$*.ActivityType = *requiredActType* **then**
7.   update *newActs$_n$* for the additional segment *aPath*.Segments$_i$
8.  **else**
9.   *newActs$_{n+1}$* $\leftarrow$ new Activity(*requiredActType*)
10.   CreateTimeConstraint(*newActs$_{n+1}$*, STARTS, interval = [0, 02:00], AFTER, *newActs$_n$*, ENDS)
11.   $n{+}{+}$
12.  update $S'$ to reflect effects of activity *newActs$_n$*
13. generate any resupply activities required
14. **return** *newActs*

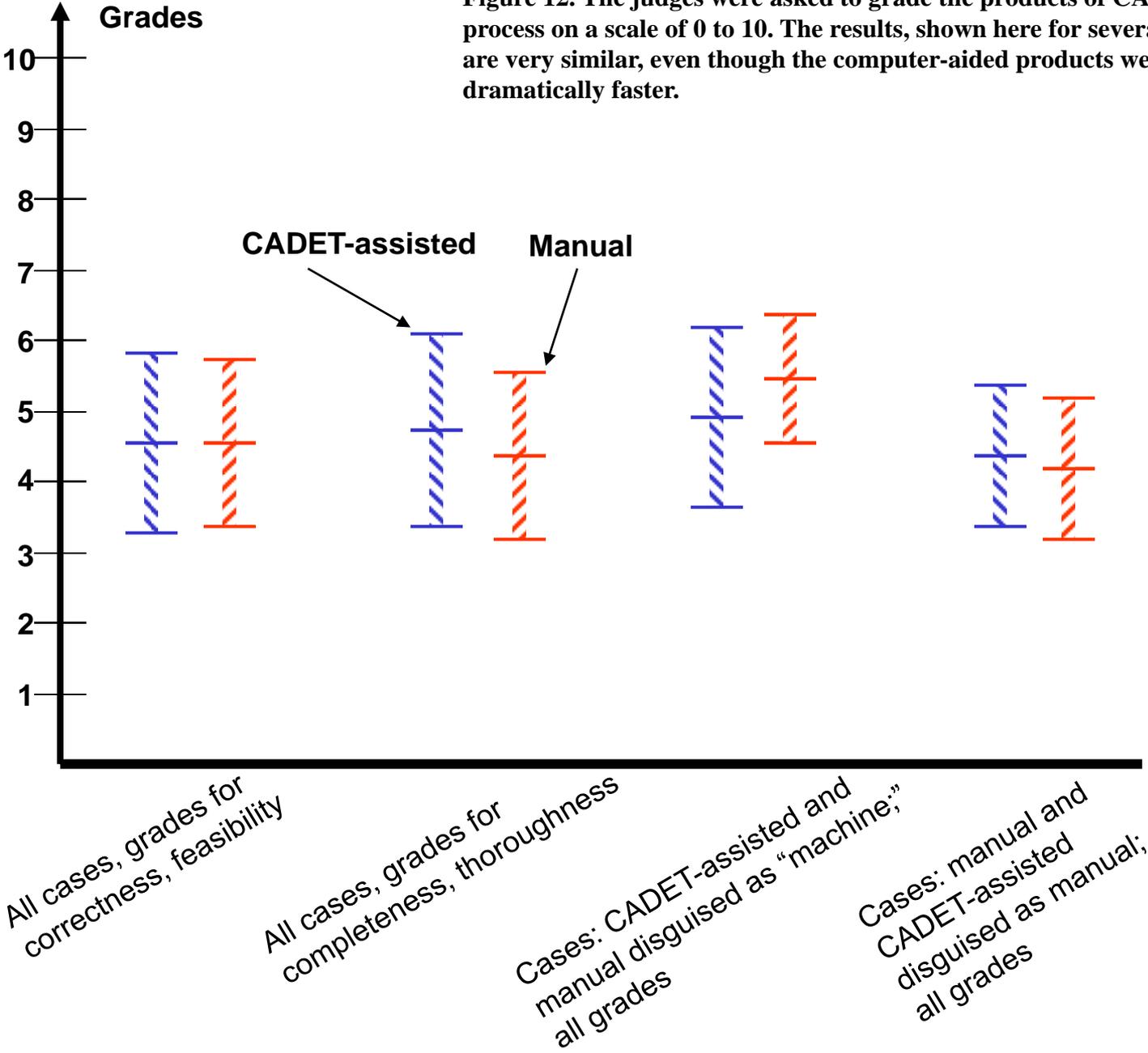

Figure 12. The judges were asked to grade the products of CADET and manual process on a scale of 0 to 10. The results, shown here for several subsets of products, are very similar, even though the computer-aided products were produced dramatically faster.

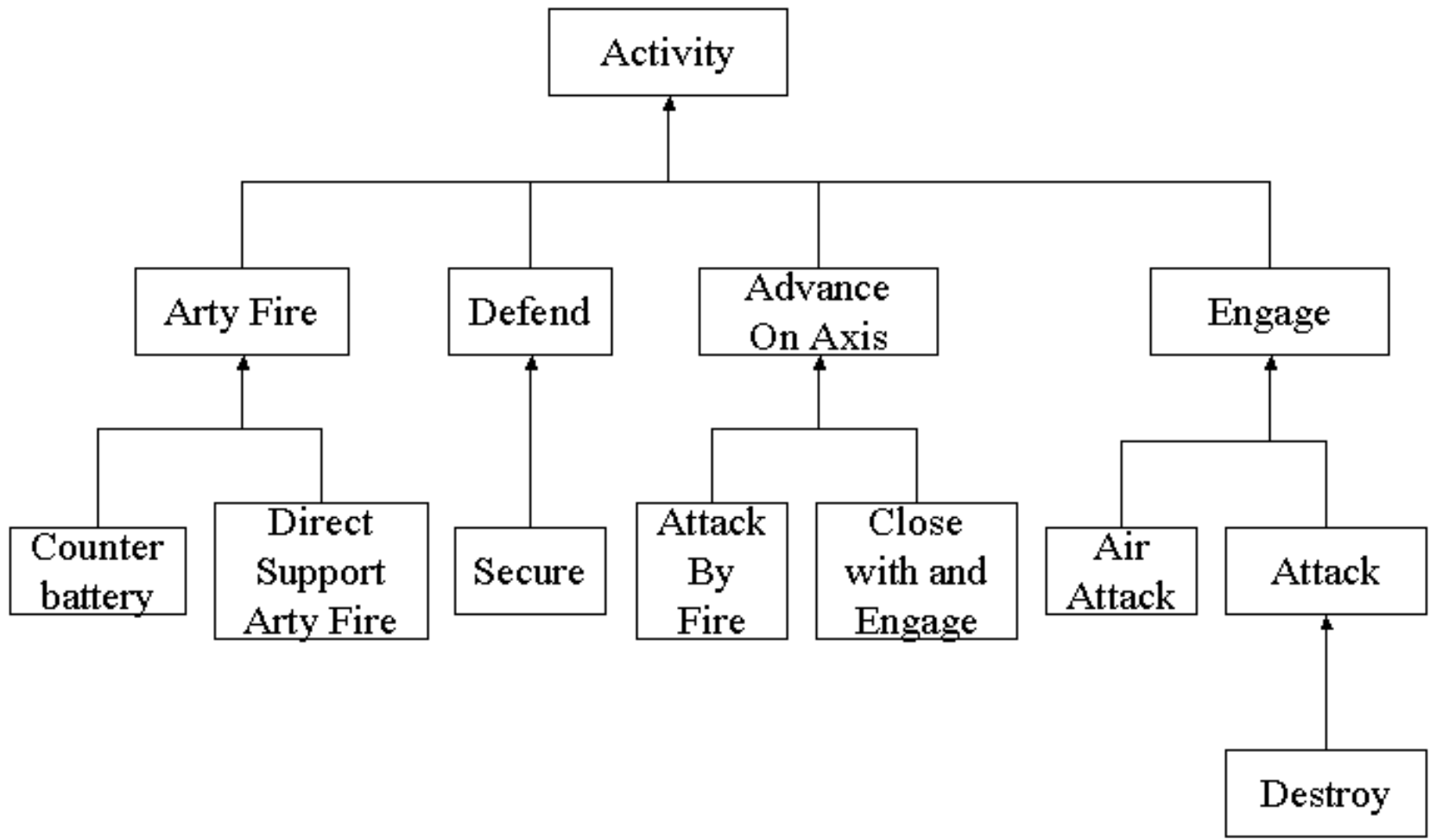

**Figure 13. A partial representation of the activity model**